\documentclass[journal]{IEEEtran}  
\pdfoutput=1
\usepackage{url}
\usepackage{graphicx}
\usepackage{makeidx}  
\begin{document}
\title{Towards Benchmarking\\ Scene Background Initialization}

\author{Lucia~Maddalena
and~Alfredo~Petrosino%
\thanks{L. Maddalena is with the National Research Council, Institute for High-Performance Computing and
Networking, 
Naples, Italy. 
e-mail: lucia.maddalena@cnr.it.}%
\thanks{A. Petrosino is with the University of Naples Parthenope,
Department of Science and Technology, Naples, 
Italy. e-mail: alfredo.petrosino@uniparthenope.it.}%
}
\maketitle              

\begin{abstract}
Given a set of images of a scene taken at different times, the
availability of an initial background model that describes the
scene without foreground objects is the prerequisite 
for a wide range of applications, ranging from video surveillance
to computational photography.
Even though several methods have been proposed for scene
background initialization, the lack of a common groundtruthed
dataset
and of a common set of metrics makes it difficult to compare their
performance.
To move first steps towards an easy and fair comparison of these
methods, we assembled a dataset of sequences frequently adopted
for background initialization, selected or created ground truths
for quantitative evaluation through a selected suite of metrics,
and compared results obtained by some existing methods, making
all the material publicly available.
\end{abstract}


\section{Introduction}

The scene background modeling process is characterized by three
main tasks: 1) \emph{model representation}, that describes the
kind of model used to represent the background; 2) \emph{model
initialization}, that regards the initialization of this model;
and 3) \emph{model update}, that concerns the mechanism used for
adapting the model to background changes along the sequence.
These tasks have been addressed by several methods, as
acknowledged by several surveys (e.g.,
\cite{Bouwmans2014,Elhabian2008}). However, most of these methods
focus on the representation and the update issues, whereas limited
attention is given to the model initialization.
The problem of scene background initialization is of interest for
a very vast audience, due to its wide range of application areas.
Indeed, the availability of an initial background model that
describes the scene without foreground objects is the
prerequisite, or at least can be of help, for many applications,
including video surveillance, video segmentation, video
compression, video inpainting, privacy protection for videos, and
computational photography (see \cite{MaddalenaChapter2014}).

We state the general problem of \emph{background initialization},
also known as {bootstrapping}, 
{background estimation}, 
{background reconstruction}, 
{initial background extraction}, 
or {background generation}, 
as follows: 
%
%
\begin{quotation}
\emph{Given a set of images of a scene taken at different times, 
in which the background is occluded by any number of foreground
objects, the aim is to determine a model
describing the scene background 
with no foreground objects.}
\end{quotation}

Depending on the application, the set of images can consist of a
subset of initial sequence frames adopted for background training
(e.g., for video surveillance), a set of non-time sequence
photographs (e.g., for computational photography), or the entire
available sequence. In the following, this set of images will be
generally referred to as the \emph{bootstrap sequence}.

In order to move first steps towards an easy and fair comparison
of existing and future background initialization methods, we
assembled and made publicly available the SBI dataset, a set of
sequences frequently adopted for background initialization,
including ground truths for quantitative evaluation through a
selected suite of metrics, and compared results obtained by some
existing methods.

\section{Sequences}

\begin{figure*}[!t]
\centering
\begin{tabular}{ccccccc}
\includegraphics[height=1.6cm]{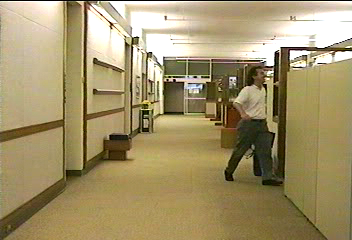} &
\includegraphics[height=1.6cm]{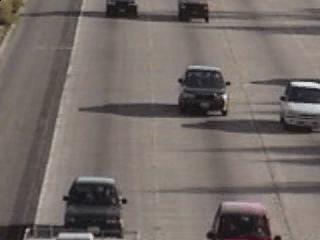} &
\includegraphics[height=1.6cm]{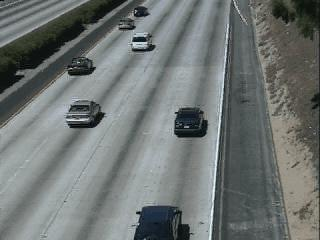} &
\includegraphics[height=1.6cm]{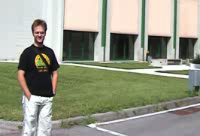} &
\includegraphics[height=1.6cm]{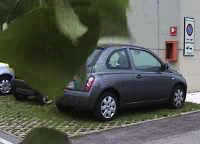} &
\includegraphics[height=1.6cm]{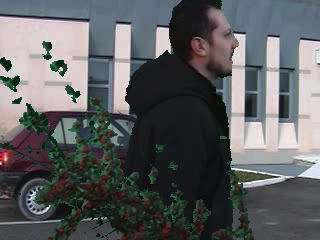} &
\includegraphics[height=1.6cm]{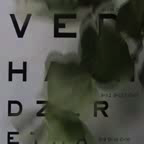} \\

frame 295 & frame 0 & frame 0 & frame 0 & frame 261 & frame 10 & frame 0\\

\includegraphics[height=1.6cm]{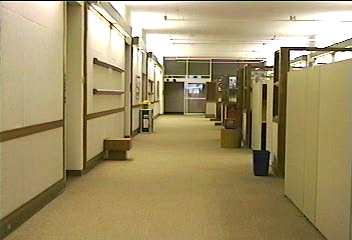} &
\includegraphics[height=1.6cm]{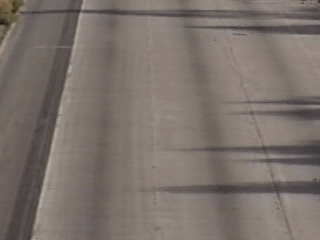} &
\includegraphics[height=1.6cm]{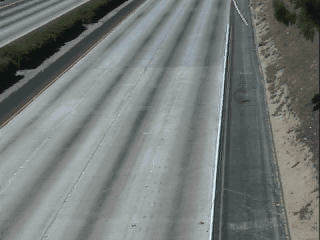} &
\includegraphics[height=1.6cm]{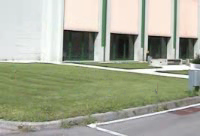} &
\includegraphics[height=1.6cm]{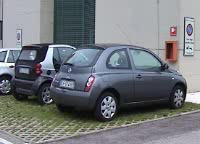} &
\includegraphics[height=1.6cm]{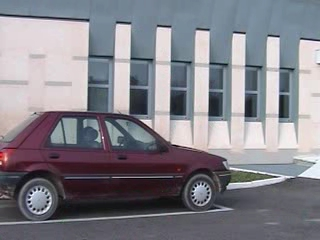} &
\includegraphics[height=1.6cm]{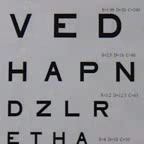}\\

{\emph{Hall\&Monitor}} & {\emph{HighwayI}} & {\emph{HighwayII}} & {\emph{CaVignal}} & {\emph{Foliage}} & {\emph{People\&Foliage}} & {\emph{Snellen}} \\
\end{tabular}
\caption{Example frames from the seven sequences of the SBI dataset (first
row) and corresponding GT (second row).} \label{figura}
\end{figure*}

The SBI dataset includes seven bootstrap sequences extracted by original publicly
available sequences that are frequently used in the literature to
evaluate background initialization algorithms; example frames are
shown in Fig. \ref{figura}. They belong to the
datasets
COST 211 (sequence \emph{Hall\&Monitor} can be found at
\url{http://www.ics.forth.gr/cvrl/demos/NEMESIS/hall_monitor.mpg}),
ATON (dataset available at
\url{http://cvrr.ucsd.edu/aton/shadow/index.html}), and
PBI (dataset available at
\url{http://www.diegm.uniud.it/fusiello/demo/bkg/}).
In Table \ref{tabella} we report, for each sequence, the name, the
dataset it belongs to, the number of available frames, the subset
of the frames adopted for testing, the original and the final
resolution. The subsets have been selected in order to avoid the
inclusion into the testing sequences of \emph{empty} frames
(frames not including foreground objects), while
the final resolution has been chosen in order to avoid problems in
the computation of boundary patches for block-based methods.
The ground truths (GT) have been manually obtained by either
choosing one of the sequence frames free of foreground objects
(not included into the subsets of used frames) or by stitching
together empty background regions from different sequence frames.
Both the complete SBI dataset and the ground truth reference
background images were made publicly available through the
SBMI2015 website at \url{http://sbmi2015.na.icar.cnr.it}.
\begin{table}[!t]
\setlength{\tabcolsep}{1.7pt} \caption{Information on sequences
adopted for evaluation.} \label{tabella} \centering
\begin{tabular}{|l|c|c|c|c|c|} \hline
                &               & \textbf{Original} & \textbf{Used}     & \textbf{Original}         & \textbf{Final}\\
 \textbf{Name}  &  \textbf{Dataset} & \textbf{frames}       & \textbf{frames}   &  \textbf{Resolution}  & \textbf{Resolution}\\\hline

\emph{Hall\&Monitor}     & COST 211  & 0-299 & 4-299 & 352x240   & 352x240\\\hline
\emph{HighwayI}          & ATON      & 0-439 & 0-439 & 320x240   & 320x240\\\hline
\emph{HighwayII}         & ATON      & 0-499 & 0-499 & 320x240   & 320x240\\\hline
\emph{CaVignal}          & PBI       & 0-257 & 0-257 & 200x136   & 200x136\\\hline
\emph{Foliage}           & PBI       & 0-399 & 6-399 & 200x148   & 200x144\\\hline
\emph{People\&Foliage}   & PBI       & 0-349 & 0-340 & 320x240   & 320x240\\\hline
\emph{Snellen}           & PBI       & 0-333 & 0-320 & 146x150   & 144x144\\\hline
\end{tabular}
\end{table}

\section{Metrics} \label{sec:Metrics}

The metrics adopted to evaluate the accuracy of the estimated
background models have been chosen among those used in the
literature for background estimation.
Denoting with $GT$ (Ground Truth) an image containing the
\emph{true} background and with $CB$ (Computed Background) the
estimated background image computed with one of the background
initialization methods, the eight adopted metrics are:
\begin{enumerate}

\item \textbf{Average Gray-level Error} (\textbf{AGE}): It is the average of the gray-level
absolute difference between $GT$ and $CB$ images. Its values range
in [0, $L$-1], where $L$ is the maximum number of grey levels;
the lower the AGE value, the better is the background estimate.

\item \textbf{Total number of Error Pixels} (\textbf{EPs}):
An \emph{error pixel} is a pixel of $CB$ whose value differs from
the value of the corresponding pixel in $GT$ by more than some
threshold $\tau$ (in the experiments the suggested value $\tau$=20 has been adopted). EPs
assume values in [$0, N$], where $N$ is the number of image
pixels; the lower the EPs value, the better is the background
estimate.

\item \textbf{Percentage of Error Pixels} (\textbf{pEPs}): It is the ratio between
the EPs and the number $N$ of image pixels. Its values range in
[0, 1]; the lower the pEPs value, the better is the background
estimate.

\item \textbf{Total number of Clustered Error Pixels} (\textbf{CEPs}):
A \emph{clustered error pixel} is defined as any error pixel whose
4-connected neighbors are also error pixels.
CEPs values range in [0, $N$]; the lower the CEPs value, the
better is the background estimate.
\item \textbf{Percentage of Clustered Error Pixels} (\textbf{pCEPs}):
It is the ratio between the CEPs and the number $N$  of image
pixels. Its values range in [0,1]; the lower the pCEPs value, the
better is the background estimate.

\item \textbf{Peak-Signal-to-Noise-Ratio} (\textbf{PSNR}):
It is defined as
$ PSNR = 10 \cdot \log_{10} \left( (L-1)^2/MSE \right),$
where $L$ is the maximum number of
grey levels and MSE is the Mean Squared Error between $GT$ and
$CB$ images.
This frequently adopted metric
assumes values in decibels (db); the higher the PSNR value, the
better is the background estimate.

\item \textbf{MultiScale Structural Similarity Index} (\textbf{MS-SSIM}):
This is the metric defined in \cite{Wang2003}, that uses
structural distortion as an estimate of the perceived visual
distortion.
It assumes values in $[0,1]$; the higher the value of $MS-SSIM$, the
better is the estimated background.

\item \textbf{Color image Quality Measure} (\textbf{CQM}):
It is a recently proposed metric \cite{Yalman2013}, based on a
reversible transformation of the YUV color space and on the PSNR
computed in the single YUV bands.
It assumes values in db and the higher the CQM value, the better
is the background estimate.

\end{enumerate}

While the last metric is defined only for color images, metrics 1)
through 7) are expressly defined for gray-scale images. In the
case of color images, they are generally applied to either the
gray-scale converted image or the luminance component Y of a color
space such as YCbCr. The latter approach has been chosen for
measurements reported in \S \ref{sec:Results}.

Matlab scripts for computing the chosen metrics were made publicly
available through the SBMI2015 website at
\url{http://sbmi2015.na.icar.cnr.it}.

\section{Experimental Results and Comparisons} \label{sec:Results}

\subsection{Compared Methods} \label{sec:Methods}

Several background initialization methods have been proposed in
the literature, as recently reviewed in
\cite{MaddalenaChapter2014}. In this study, we compared five of
them, based on different methodological schemes.

The method considered here as the baseline method is the temporal
{\bf Median}, that computes the value of each background pixel as
the median of pixel values at the same location throughout the
whole bootstrap sequence (e.g.,
\cite{Gloyer1995,Maddalena3dSOBS+2014}).
In the reported experiments on color bootstrap sequences, the
temporal median is computed for each pixel as the one that
minimizes the sum of $L_\infty$ distances of the pixel from all
the other pixels.

%
The Self-Organizing Background Subtraction (SOBS) algorithm
\cite{Maddalena2008} and its spatially coherent extension {\bf
SC-SOBS} \cite{Maddalena2012} implement an approach to moving
object detection based on the neural background model
automatically generated by a self-organizing method without prior
knowledge about the involved patterns.
For each pixel, the neuronal map consists of $n \times n$ weight
vectors, each initialized with the pixel value. The whole set of
weight vectors for all pixels is organized as a 2D neuronal map
topologically organized such that adjacent blocks of $n \times n$
weight vectors model corresponding adjacent pixels in the image.
Even though not explicitly devoted to background initialization,
the method has been chosen as an example of method based on
temporal statistics. Indeed, the first learning phase (usually
followed by an on-line phase for moving object detection),
provides an initial estimate of the background, obtained through a
selective update procedure over the bootstrap sequence,
taking into account spatial coherence.
In the experiments, the background estimate is obtained as the
result of the initial training of the software SC-SOBS (publicly
available in the download section of the CVPRLab at
\url{http://cvprlab.uniparthenope.it}) using for all the sequences
the same default parameter values. Once the neural background
model is computed, the background estimate is extracted for each
pixel by choosing, among the $n^2$ modeling weight vectors, the
one that is closest to the ground truth. Indeed, this method
provides the best representation of the background that can be
achieved by SC-SOBS, even though it is only applicable for
comparison purposes, being based on the existence of a ground
truth to compare with.

%
%
The pixel-level, non-recursive method based on subsequences of
stable intensity proposed in \cite{Wang2006} (in the following
denoted as {\bf WS2006}) employs a two-phase approach. Relying on
the assumption that a background value always has the longest
stable value, for each pixel (or image block) different
non-overlapping temporal subsequences with similar intensity
values (``stable subsequences'') are first selected. The most
reliable subsequence, which is more likely to arise from the
background, is thenchosen based on the RANSAC method. The temporal
mean of the selected subsequence provides the estimated background
model.
For the reported experiments, WS2006 has been implemented based on
\cite{Wang2006}, and parameter values have been chosen among those
suggested by the authors and providing the best overall results.

%
In the block-level, recursive, iterative model completion
technique proposed in \cite{Reddy2011} (in the following denoted
as {\bf RSL2011}), for each block location of the bootstrap
sequence, a representative set of distinct blocks is maintained
along its temporal line. The background estimation is carried out
in a Markov Random Field framework, where the clique potentials
are computed based on the combined frequency response of the
candidate block and its neighborhood.
Spatial continuity of structures within a scene is enforced by the
assumption that the most appropriate block provides the smoothest
response.
The reported experimental results have been obtained through the
related software publicly available at
\url{http://arma.sourceforge.net/background_est/}.

{\bf Photomontage}
%
%
provides an example of method for background initialization
approach\-ed as optimal labeling \cite{Agarwala2004}. It is an
unified framework  for interactive image composition,  based on
energy minimization, under which various image editing tasks can
be done by choosing appropriate energy functions.
The cost function, minimized through graph cuts, consists of an
{interaction term}, that penalizes perceivable seams in the
composite image, and a {data term}, that reflects various
objectives of different image editing tasks. For the specific task
of background estimation, the data term adopted for achieving
visual smoothness is the maximum likelihood image objective.
The reported experimental results have been obtained through the
related software publicly available at
\url{http://grail.cs.washington.edu/projects/photomontage/}.

\subsection{Qualitative and Quantitative Evaluation}

\begin{figure*}[!t]
\centering
\begin{tabular}{ccccccc}
\rotatebox{90}{\makebox[1.3cm][c]{\emph{\tiny Hall\&Monitor}}} &
\includegraphics[width=.13\textwidth]{Images/GT/GT_HallAndMonitor.png} &
\includegraphics[width=.13\textwidth]{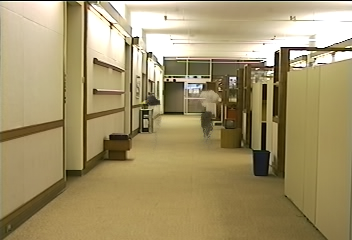} &
\includegraphics[width=.13\textwidth]{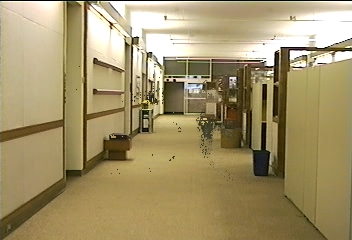} &
\includegraphics[width=.13\textwidth]{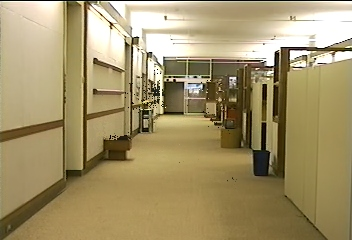}  &
\includegraphics[width=.13\textwidth]{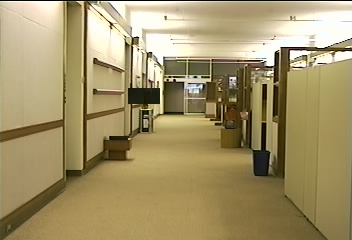} &
\includegraphics[width=.13\textwidth]{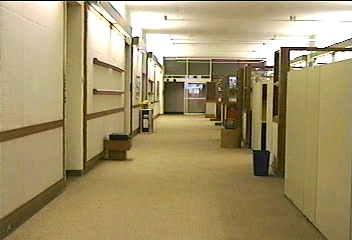}  \\

\rotatebox{90}{\makebox[1.3cm][c]{\emph{\tiny HighwayI}}} &
\includegraphics[width=.13\textwidth]{Images/GT/GT_HighwayI.png} &
\includegraphics[width=.13\textwidth]{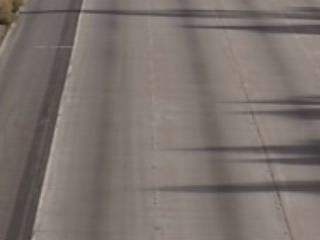} &
\includegraphics[width=.13\textwidth]{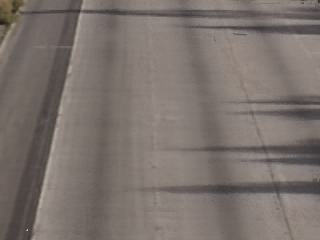} &
\includegraphics[width=.13\textwidth]{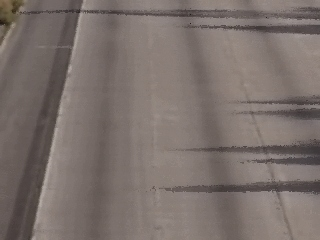} &
\includegraphics[width=.13\textwidth]{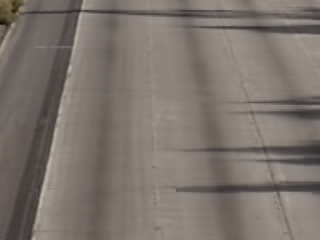} &
\includegraphics[width=.13\textwidth]{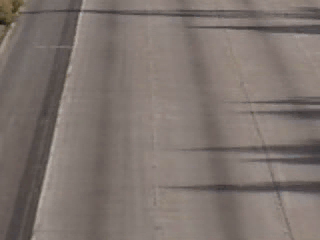}  \\

\rotatebox{90}{\makebox[1.5cm][c]{\emph{\tiny HighwayII}}} &
\includegraphics[width=.13\textwidth]{Images/GT/GT_HighwayII.png} &
\includegraphics[width=.13\textwidth]{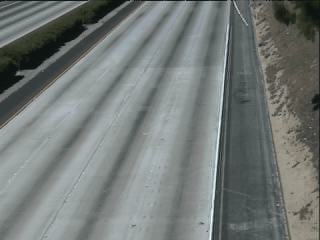} &
\includegraphics[width=.13\textwidth]{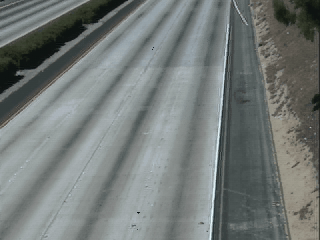} &
\includegraphics[width=.13\textwidth]{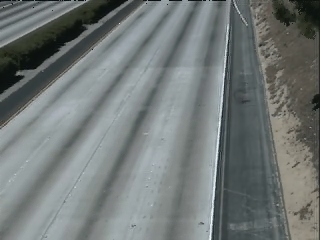} &
\includegraphics[width=.13\textwidth]{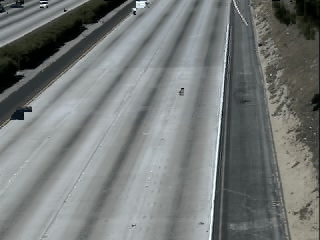} &
\includegraphics[width=.13\textwidth]{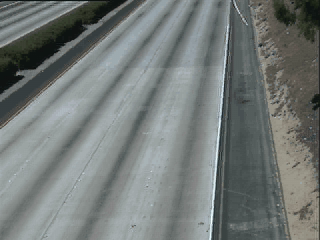} \\

\rotatebox{90}{\makebox[1.3cm][c]{\emph{\tiny CaVignal}}} &
\includegraphics[width=.13\textwidth]{Images/GT/GT_CaVignal.png} &
\includegraphics[width=.13\textwidth]{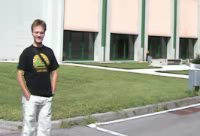} &
\includegraphics[width=.13\textwidth]{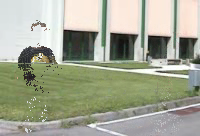} &
\includegraphics[width=.13\textwidth]{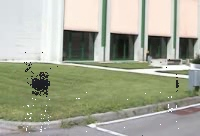} &
\includegraphics[width=.13\textwidth]{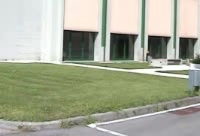} &
\includegraphics[width=.13\textwidth]{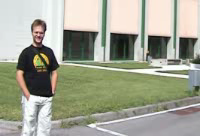}  \\

\rotatebox{90}{\makebox[1.3cm][c]{\emph{\tiny Foliage}}} &
\includegraphics[width=.13\textwidth]{Images/GT/GT_Foliage.png} &
\includegraphics[width=.13\textwidth]{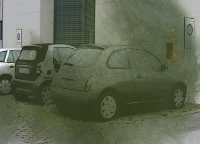} &
\includegraphics[width=.13\textwidth]{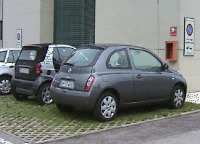} &
\includegraphics[width=.13\textwidth]{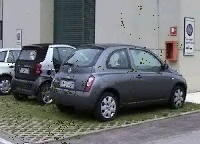} &
\includegraphics[width=.13\textwidth]{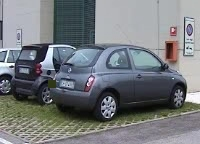} &
\includegraphics[width=.13\textwidth]{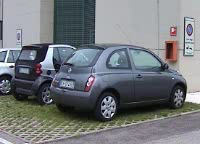}  \\

\rotatebox{90}{\makebox[1.4cm][c]{\emph{\tiny People\&Foliage}}} &
\includegraphics[width=.13\textwidth]{Images/GT/GT_PeopleAndFoliage.png} &
\includegraphics[width=.13\textwidth]{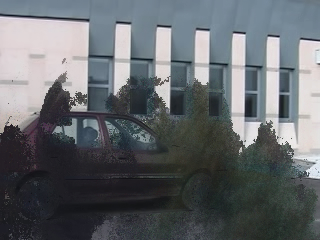} &
\includegraphics[width=.13\textwidth]{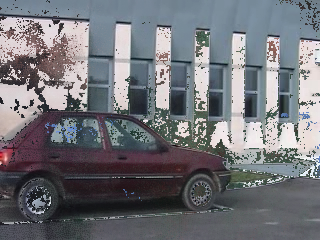} &
\includegraphics[width=.13\textwidth]{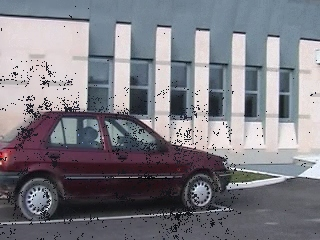} &
\includegraphics[width=.13\textwidth]{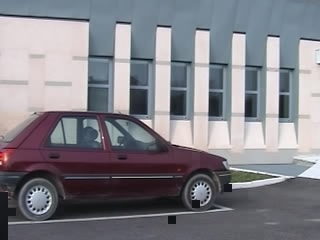} &
\includegraphics[width=.13\textwidth]{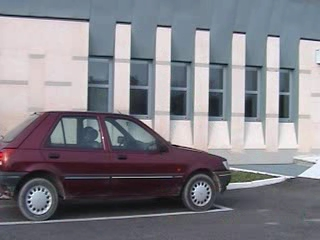}  \\

\rotatebox{90}{\makebox[1.7cm][c]{\emph{\tiny Snellen}}} &
\includegraphics[width=.13\textwidth]{Images/GT/GT_Snellen.png} &
\includegraphics[width=.13\textwidth]{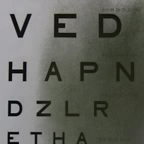} &
\includegraphics[width=.13\textwidth]{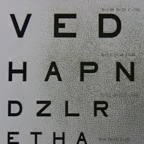} &
\includegraphics[width=.13\textwidth]{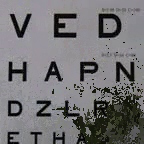} &
\includegraphics[width=.13\textwidth]{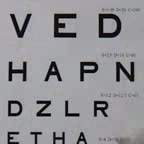} &
\includegraphics[width=.13\textwidth]{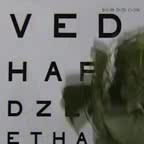}  \\

 & (a) & (b) & (c) & (d) & (e) & (f)\\
\end{tabular}

\caption{Comparison of background initialization results on the
SBI dataset obtained by: (a) GT, (b) Median, (c) SC-SOBS, (d)
WS2006, (e) RSL2011, and (f) Photomontage} \label{fig:AllSeq}
\end{figure*}

%
In Fig. \ref{fig:AllSeq} we show the background images obtained
by the compared methods on the SBI dataset, while in Table
\ref{table:AllSeq} we report accuracy results according to the
metrics described in \S \ref{sec:Metrics}.

For sequence \emph{Hall\&Monitor}, we observe few differences in
initializing the background in image regions where foreground
objects are more persistent during the sequence. A man walking
straight down the corridor occupies the same image region for more
than 65\% of the sequence frames, while the briefcase is left on
the small table for the last 60\% of sequence frames. Only WS2006,
RSL2011, and Photomontage well handle the walking man, but they
include the abandoned briefcase into the background.
This qualitative analysis is confirmed by accuracy results in
terms of EPs and CEPs values reported in Table \ref{table:AllSeq}.
Moreover, AGE values are quite low for all the compared methods,
due to the reduced size of foreground objects as compared to the
image size. However, the worst AGE values are achieved by RSL2011
and Photomontage, despite their quite good qualitative results.
Finally, all the compared methods achieve similar values of PSNR,
MS-SSIM, and CQM, as overall, apart from reduced sized defects
related to foreground objects, they all succeed in providing a
sufficiently faithful representation of the empty background.
%
%

For both \emph{HighwayI} and \emph{HighwayII} sequences, all the
compared methods succeed in providing a faithful representation of
the background model. This is due to the fact that, even though
the highway is always fairly crowded by passing cars, the
background is revealed for at least 50\% of the entire bootstrap
sequence length and no cars remain stationary during the sequence.
The above qualitative considerations are only partially confirmed
by performance results reported in Table \ref{table:AllSeq}.
Indeed, different AGE and EPs values are achieved by qualitatively
similar estimated backgrounds, while similar low CEPs values and
high MS-SSIM, PSNR, and CQM values are achieved by all the
compared methods.
\begin{table*}[!t]
\caption{Accuracy results of the
compared methods on the SBI dataset.} \label{table:AllSeq}
\centering
\begin{tabular}{|l|c|c|c|c|c|c|c|c|c|} \hline
\multicolumn{9}{|c|}{\emph{Hall\&Monitor}}\\\hline
\textbf{Method} & \textbf{AGE} & \textbf{EPs}   & \textbf{pEPs}   & \textbf{CEPs} & \textbf{pCEPs}  & \textbf{MS-SSIM}   & \textbf{PSNR}  &  \textbf{CQM}\\\hline
Median          & 2.7105 &    839 & 0.9931\% &    451 & 0.5339\% & 0.9640 & 30.4656 & 42.6705 \\\hline
SC-SOBS         & 2.4493 &    828 & 0.9801\% &    272 & 0.3220\% & 0.9653 & 30.4384 & 43.1867 \\\hline
WS2006          & 2.6644 &    470 & 0.5563\% &     26 & 0.0308\% & 0.9821 & 30.9313 & 40.0949 \\\hline
RSL2011         & 3.2687 &    703 & 0.8321\% &    398 & 0.4711\% & 0.9584 & 28.4428 & 37.9971 \\\hline
Photomontage    & 2.7986 &    305 & 0.3610\% &     69 & 0.0817\% & 0.9819 & 33.3715 & 41.7323 \\\hline

\multicolumn{9}{|c|}{\emph{HighwayI}}\\\hline
Median       & 1.4275 &    120 & 0.1563\% &     11 & 0.0143\% & 0.9924 & 40.1432 & 62.5723 \\\hline
SC-SOBS      & 1.2286 &      3 & 0.0039\% &      0 & 0.0000\% & 0.9949 & 42.6868 & 65.5755 \\\hline
WS2006       & 2.5185 &    526 & 0.6849\% &     19 & 0.0247\% & 0.9816 & 35.6885 & 56.9113 \\\hline
RSL2011      & 2.8139 &    267 & 0.3477\% &     33 & 0.0430\% & 0.9830 & 36.0290 & 51.9835 \\\hline
Photomontage & 2.1745 &    313 & 0.4076\% &     37 & 0.0482\% & 0.9830 & 37.1250 & 59.0270 \\\hline

\multicolumn{9}{|c|}{\emph{HighwayII}}\\\hline
Median       & 1.7278 &    245 & 0.3190\% &      1 & 0.0013\% & 0.9961 & 34.6639 & 42.3162 \\\hline
SC-SOBS      & 0.6536 &      7 & 0.0091\% &      0 & 0.0000\% & 0.9982 & 44.6312 & 54.3785 \\\hline
WS2006       & 2.4906 &    375 & 0.4883\% &     10 & 0.0130\% & 0.9927 & 33.9515 & 40.5088 \\\hline
RSL2011      & 5.6807 &    956 & 1.2448\% &    316 & 0.4115\% & 0.9766 & 28.6703 & 35.0821 \\\hline
Photomontage & 2.4306 &    452 & 0.5885\% &      4 & 0.0052\% & 0.9909 & 34.3975 & 41.7656 \\\hline

\multicolumn{9}{|c|}{\emph{CaVignal}}\\\hline
Median       & 10.3082 &   2846 & 10.4632\% &   2205 & 8.1066\% & 0.7984 & 18.1355 & 33.1438 \\\hline
SC-SOBS      & 4.0941 &    869 & 3.1949\% &    436 & 1.6029\% & 0.8779 & 21.8507 & 42.2652 \\\hline
WS2006       & 2.5403 &    408 & 1.5000\% &    129 & 0.4743\% & 0.9289 & 27.1089 & 37.0609 \\\hline
RSL2011      & 1.6132 &      4 & 0.0147\% &      0 & 0.0000\% & 0.9967 & 41.3795 & 52.5856 \\\hline
Photomontage & 11.2665 &   3052 & 11.2206\% &   2408 & 8.8529\% & 0.7919 & 17.6257 & 32.0570 \\\hline

\multicolumn{9}{|c|}{\emph{Foliage}}\\\hline
Median       & 27.0135 &  13626 & 47.3125\% &   8772 & 30.4583\% & 0.6444 & 16.7842 & 28.7321 \\\hline
SC-SOBS      & 3.8215 &    160 & 0.5556\% &      0 & 0.0000\% & 0.9900 & 31.7713 & 39.1387 \\\hline
WS2006       & 6.8649 &    821 & 2.8507\% &      2 & 0.0069\% & 0.9754 & 27.2438 & 34.9776 \\\hline
RSL2011      & 2.2773 &     43 & 0.1493\% &     11 & 0.0382\% & 0.9951 & 36.7450 & 43.1208 \\\hline
Photomontage & 1.8592 &      0 & 0.0000\% &      0 & 0.0000\% & 0.9974 & 39.1779 & 45.6052 \\\hline

\multicolumn{9}{|c|}{\emph{People\&Foliage}}\\\hline
Median       & 24.4211 &  24760 & 32.2396\% &  19446 & 25.3203\% & 0.6114 & 15.1870 & 27.4979 \\\hline
SC-SOBS      & 15.1031 &  10770 & 14.0234\% &   3849 & 5.0117\% & 0.7561 & 16.6189 & 35.3667 \\\hline
WS2006       & 5.4243 &   2743 & 3.5716\% &     71 & 0.0924\% & 0.9269 & 22.6952 & 31.3847 \\\hline
RSL2011      & 2.0980 &    612 & 0.7969\% &    434 & 0.5651\% & 0.9905 & 32.5550 & 37.0598 \\\hline
Photomontage & 1.4103 &      3 & 0.0039\% &      0 & 0.0000\% & 0.9973 & 41.0866 & 47.1517 \\\hline

\multicolumn{9}{|c|}{\emph{Snellen}}\\\hline
Median       & 42.3981 &  12898 & 62.2010\% &  11814 & 56.9734\% & 0.6932 & 13.6573 & 36.0691 \\\hline
SC-SOBS      & 16.8898 &   7746 & 37.3553\% &   5055 & 24.3779\% & 0.9303 & 21.2571 & 44.7498 \\\hline
WS2006       & 23.0010 &   4804 & 23.1674\% &   2544 & 12.2685\% & 0.7481 & 15.6158 & 24.9930 \\\hline
RSL2011      & 1.8095 &    133 & 0.6414\% &     99 & 0.4774\% & 0.9979 & 38.0295 & 50.2600 \\\hline
Photomontage & 29.9797 &   6946 & 33.4973\% &   6318 & 30.4688\% & 0.5926 & 14.1466 & 26.9210 \\\hline

\end{tabular}
\end{table*}

Sequence \emph{CaVignal} represents a major burden for most of the
compared methods. Indeed, the only man appearing in the sequence
stands still on the left of the scene for the first 60\% of
sequence frames; then starts walking and rests on the right of the
scene for the last 10\% of sequence frames.
The persistent clutter at the beginning of the scene leads most of
the compared methods to include the man on the left into the
estimated background, while the persistent clutter at the end of
the scene leads only WS2006 to partially include the man on the
right into the background. Only RSL2011 perfectly handles the
persistent clutter, accordingly achieving the best accuracy
results for all the metrics.
%

For sequence \emph{Foliage}, even though moving leaves occupy most
of the background area for most of the time, many of the compared
methods achieve a quite good representation of the scene
background.
Indeed, only Median produces a greenish halo due to the foreground
leaves over almost the entire scene area, and accordingly achieves
the worst accuracy results for all the metrics.
%

Also sequence \emph{People\&Foliage} is problematic for most of
the compared methods. Indeed, the artificially added leaves and
men occupy almost all the scene area in almost all the sequence
frames.
Only Photomontage and RSL2011 appear to well handle the wide
clutter, also achieving the best accuracy results for all the
metrics.
%

In sequence \emph{Snellen}, the foreground leaves occupy almost
all the scene area in almost all the sequence frames. This leads
most of the methods to include the contribution of leaves into the
final background model. The best qualitative result can be
attributed to RSL2011, as confirmed by the quantitative analysis
in terms of all the adopted metrics.

Overall, we can observe that
most of the best performing background initialization methods are
region-based or hybrid, confirming the importance of taking into
account spatio-temporal inter-pixel relations.
Also selectivity in choosing the best candidate pixels, shared by
all the best performing methods, appears to be important for
achieving accurate results.
Instead, all the common methodological schemes shared by the
compared methods can lead to accurate results, showing no
preferred scheme, and the same can be said concerning recursivity.

In order to assess the challenge that each sequence poses for the
tested methods, we further
computed the median values of all metrics obtained by the compared
methods for each sequence, and
ranked the sequences according to these median values, as shown in
Table \ref{table:ClassificaSequenze}.
Here, \emph{HighwayI}  and \emph{HighwayII} sequences reveal as
those that are best handled by all methods (in the sense of
median), while \emph{Snellen} is the worst handled.
Bearing in mind the kind of foreground objects included into the
sequences, we can observe that their size is not a major burden;
e.g., \emph{Foliage} sequence is better handled than
\emph{Hall\&Monitor}, even though the size of the foreground
objects is much larger. Instead, their speed (or their steadiness)
has much greater influence on the results. As instance,
\emph{CaVignal} sequence is worse handled than \emph{Foliage}, 
since it includes almost static foreground objects that are
frequently misinterpreted as background.
It can also be observed that the median values of pEPs and MS-SSIM
metrics perfectly vary according to the difficulty in handling the
sequences; these two metrics confirm to be strongly indicative of
the performance of background initialization methods.

\begin{table*}[!t]
\caption{Median values of all
metrics obtained by the compared methods for each sequence and
average rank of the sequences according to these median values.}
\label{table:ClassificaSequenze} \centering
\begin{tabular}{|l|c|c|c|c|c|c|c|c|c|} \hline
\textbf{Sequence}    & \textbf{Av.}  &
\textbf{Median} & \textbf{Median}   & \textbf{Median}
& \textbf{Median} & \textbf{Median}  & \textbf{Median} & \textbf{Median}  &  \textbf{Median}\\

  & \textbf{rank}  &
\textbf{AGE} & \textbf{EPs}   & \textbf{pEPs}
& \textbf{CEPs} & \textbf{pCEPs}  & \textbf{MS-SSIM} & \textbf{PSNR}  &  \textbf{CQM}\\\hline

HighwayI        & 1,63 & 2,17 & 267 & 0,35\% & 19  & 0,02\% & 0,97 & 37,13 & 59,03\\\hline
HighwayII       & 2,00 & 2,43 & 375 & 0,49\% & 4   & 0,01\% & 0,95 & 34,40 & 41,77\\\hline
Foliage         & 2,50 & 3,82 & 160 & 0,56\% & 2   & 0,01\% & 0,95 & 31,77 & 39,14\\\hline
Hall\&Monitor   & 3,75 & 2,71 & 703 & 0,83\% & 272 & 0,32\% & 0,95 & 30,47 & 41,73\\\hline
CaVignal        & 5,63 & 4,09 & 956 & 3,19\% & 435 & 1,60\% & 0,89 & 21,85 & 35,08\\\hline
People\&Foliage & 5,63 & 5,42 & 2743& 3,57\% & 434 & 0,57\% & 0,78 & 22,70 & 35,37\\\hline
Snellen         & 6,75 & 23,00& 6946&33,50\% & 5055& 24,38\%& 0,76 & 15,62 & 36,07\\\hline
\end{tabular}
\end{table*}

\section{Concluding Remarks}  \label{sec:Conclusions}

We proposed a benchmarking study 
for scene background initialization, moving the first steps towards a
fair and easy comparison of existing and future methods, on a
common dataset of groundtruthed sequences, with a common set of
metrics, and based on reproducible results.
The assembled SBI dataset, the ground truths, and a tool to
compute the suite of metrics were made publicly available.

Based on the benchmarking study, first considerations have been
drawn.

Concerning main issues in background initialization, low speed (or
steadiness), rather than great size, of foreground objects
included into the bootstrap sequence is a major burden for most of
the methods.

All the common methodologies shared by the compared methods can
lead to accurate results, showing no preferred scheme, and the
same can be said concerning recursivity.
Anyway,
the best results are generally achieved by methods that are
region-based or hybrid, and selective; thus, these are the methods
to be preferred.

Another conclusion can be drawn, concerning the evaluation of
background initialization methods. Among the eight selected
metrics frequently adopted in the literature, pEPs and MS-SSIM
confirm to be strongly indicative of the performance of background
initialization methods. This can be of peculiar interest for
evaluating future background initialization methods.

\section*{Acknowledgment}

This research was supported by Project PON01\_01430 PT2LOG under
the Research and Competitiveness PON, funded by the European Union
(EU) via structural funds, with the responsibility of the Italian
Ministry of Education, University, and Research (MIUR).
%

%
%
\bibliographystyle{IEEEtran}
\bibliography{IEEEabrv,BookBibliography}

\begin{thebibliography}{10}
\providecommand{\url}[1]{#1}
\csname url@samestyle\endcsname
\providecommand{\newblock}{\relax}
\providecommand{\bibinfo}[2]{#2}
\providecommand{\BIBentrySTDinterwordspacing}{\spaceskip=0pt\relax}
\providecommand{\BIBentryALTinterwordstretchfactor}{4}
\providecommand{\BIBentryALTinterwordspacing}{\spaceskip=\fontdimen2\font plus
\BIBentryALTinterwordstretchfactor\fontdimen3\font minus
  \fontdimen4\font\relax}
\providecommand{\BIBforeignlanguage}[2]{{%
\expandafter\ifx\csname l@#1\endcsname\relax
\typeout{** WARNING: IEEEtran.bst: No hyphenation pattern has been}%
\typeout{** loaded for the language `#1'. Using the pattern for}%
\typeout{** the default language instead.}%
\else
\language=\csname l@#1\endcsname
\fi
#2}}
\providecommand{\BIBdecl}{\relax}
\BIBdecl

\bibitem{Bouwmans2014}
T.~Bouwmans, ``Traditional and recent approaches in background modeling for
  foreground detection: An overview,'' \emph{Computer Science Review}, vol.
  11–12, pp. 31--66, 2014.

\bibitem{Elhabian2008}
S.~Elhabian, K.~El~Sayed, and S.~Ahmed, ``Moving object detection in spatial
  domain using background removal techniques: State-of-art,'' \emph{Recent
  Patents on Computer Science}, vol.~1, no.~1, pp. 32--54, Jan. 2008.

\bibitem{MaddalenaChapter2014}
L.~Maddalena and A.~Petrosino, ``Background model initialization for static
  cameras,'' in \emph{Background Modeling and Foreground Detection for Video
  Surveillance}, T.~Bouwmans, F.~Porikli, B.~H�ferlin, and A.~Vacavant,
  Eds.\hskip 1em plus 0.5em minus 0.4em\relax Chapman \& Hall/CRC, 2014, pp.
  3--1--3--16.

\bibitem{Wang2003}
Z.~Wang, E.~Simoncelli, and A.~Bovik, ``Multiscale structural similarity for
  image quality assessment,'' in \emph{Signals, Systems and Computers, 2004.
  Conference Record of the Thirty-Seventh Asilomar Conference on}, vol.~2,
  2003, pp. 1398--1402 Vol.2.

\bibitem{Yalman2013}
Y.~Yalman and I.~Erturk, ``A new color image quality measure based on {YUV}
  transformation and {PSNR} for human vision system,'' \emph{Turkish J. of
  Electrical Eng. \& Comput. Sci.}, vol.~21, pp. 603--612, 2013.

\bibitem{Gloyer1995}
B.~Gloyer, H.~K. Aghajan, K.-Y. Siu, and T.~Kailath, ``Video-based
  freeway-monitoring system using recursive vehicle tracking,'' pp. 173--180,
  1995.

\bibitem{Maddalena3dSOBS+2014}
L.~Maddalena and A.~Petrosino, ``The {3dSOBS+} algorithm for moving object
  detection,'' \emph{Comput. Vis. Image Underst.}, vol. 122, pp. 65--73, 2014.

\bibitem{Maddalena2008}
------, ``A self-organizing approach to background subtraction for visual
  surveillance applications,'' \emph{{IEEE} Trans. Image Process.}, vol.~17,
  no.~7, pp. 1168--1177, July 2008.

\bibitem{Maddalena2012}
------, ``The {SOBS} algorithm: What are the limits?'' in \emph{Proc. CVPR
  Workshops}, June 2012, pp. 21--26.

\bibitem{Wang2006}
H.~Wang and D.~Suter, ``A novel robust statistical method for background
  initialization and visual surveillance,'' in \emph{Proc. ACCV'06}.\hskip 1em
  plus 0.5em minus 0.4em\relax Berlin, Heidelberg: Springer-Verlag, 2006, pp.
  328--337.

\bibitem{Reddy2011}
V.~Reddy, C.~Sanderson, and B.~C. Lovell, ``A low-complexity algorithm for
  static background estimation from cluttered image sequences in surveillance
  contexts,'' \emph{EURASIP J. Image Video Process.}, vol. 2011, pp. 1:1--1:14,
  Jan. 2011.

\bibitem{Agarwala2004}
A.~Agarwala, M.~Dontcheva, M.~Agrawala, S.~Drucker, A.~Colburn, B.~Curless,
  D.~Salesin, and M.~Cohen, ``Interactive digital photomontage,'' \emph{ACM
  Trans. Graph.}, vol.~23, no.~3, pp. 294--302, Aug. 2004.

\end{thebibliography}

\end{document}